\documentclass{article}

\usepackage[T1]{fontenc}      
\usepackage[utf8]{inputenc} 
\usepackage[english]{babel}

\usepackage{amssymb,amsthm,amsfonts,amscd}
\usepackage{amsmath}
\usepackage{times}

\usepackage[authoryear,sort&compress]{natbib}

\usepackage{makeidx} 
\usepackage{graphicx}
\usepackage{algorithm}
\usepackage{algorithmic}
\usepackage{natbib}
\graphicspath{{.},{img/}}

\newcommand \R{\mathbb R}

\makeindex

\usepackage{vmargin}
\title{The Consciousness Prior}
\author{
  Yoshua Bengio\\
  Universit\'{e} de Montr\'{e}al, Mila
  \thanks{Also CIFAR Senior Fellow}
}
\date{First posted October 15th 2017; revised, December 1, 2019}

\begin{document}

\maketitle

\abstract{A new prior is proposed for learning representations of high-level concepts of the kind we manipulate with language. This prior can be combined with other priors in order to help disentangling abstract factors from each other. It is inspired by cognitive neuroscience theories of consciousness, seen as a bottleneck through which just a few elements, after having been selected by attention from a broader pool, are then broadcast and condition further processing, both in perception and decision-making. The set of recently selected elements one becomes aware of is seen as forming a low-dimensional conscious state. This conscious state is combining the few concepts constituting a conscious thought, i.e., what one is immediately conscious of at a particular moment. We claim that this architectural and information-processing constraint corresponds to assumptions about the joint distribution between high-level concepts. To the extent that these assumptions are generally true (and the form of natural language seems consistent with them), they can form a useful prior for representation learning. A low-dimensional thought or conscious state is analogous to a sentence: it involves only a few variables and yet can make a statement with very high probability of being true. This is
consistent with a joint distribution (over high-level concepts) which has the form of a sparse factor graph, i.e., where the dependencies captured by each factor of the factor graph involve only very few variables while creating a strong dip in the overall energy function. Instead of making predictions in the sensory (e.g. pixel) space, one can thus make predictions in this high-level abstract space, which do not have to be limited to just the next time step but can relate events far away from each other in time. The consciousness prior also makes it natural to map conscious states to natural language utterances or to express classical AI knowledge in a form similar to facts and rules, albeit capturing uncertainty as well as efficient search mechanisms implemented by attention mechanisms.  }


\section{Introduction}

We propose here a new kind of prior for top-level abstract representations of concepts of the kind
humans manipulate with natural language,
inspired by modern theories of consciousness such as the global workspace theory~\citep{Baars-1988,Baars-1997,Dehaene+Naccache-2001,Baars-2002,Dehaene-et-al-2017}
as a form of awareness~\citep{vanGulick-2004},
i.e., as defined by Locke, consciousness is ``the perception of what passes
in a man’s own mind'', or awareness of an external object or
something within oneself (Wikipedia definition). The main contribution
of this paper is proposing a machine learning justification for an aspect
of this theory, stipulating that elements of a conscious thought are selected through an attention mechanism (such as the content-based attention
mechanism we introduced in~\citep{Bahdanau-et-al-ICLR2015-small}) and then broadcast to the rest of the brain, strongly
influencing downstream perception and action as well as the content of the
next conscious thought. The paper sees this as a computational mechanism
which is consistent with a hypothesis about the form of the joint distribution
between the type of high-level variables which can form a conscious thought.
Since a conscious thought only refers to very few variables at a time,
we suggest that this corresponds to a form of knowledge representation
which is factored into pieces involving a few variables at a time.
From a probabilistic modeling point of view, this corresponds to
a {\em sparse factor graph}. Each ``factor" captures the possibly strong
dependency between a few variables. Although a variable can participate
in many such factors, each factor links very few variables, similarly to
words or concepts linked together in a sentence in natural language.

\section{System 2 Processing and Global Workspace Theory of Consciousness}

For lack of a generally accepted definition of consciousness - because there are still many competing theories - we consider conscious aspects
of cognition as those which humans can report about through language.
We closely associate conscious processing to Kahneman's system 2 cognitive abilities~\citep{Kahneman-2011}. 
System 1 tasks align well with the current
successful applications of deep learning, e.g., low-level
perception (and to a lesser extent low-level action)
and intuitive knowledge (e.g. knowing that a particular
Go move is good or that a given picture contains
the image of a dog), i.e., knowledge which is difficult
to verbalize, and which can typically be applied very
quickly (in less than a second). On the other hand,
system 2 cognitive abilities are those which can
can be described verbally, and thus includes the part of
our cognitive abilities which we can communicate explicitly
to a computer (typically as a sequence of computational
steps), and include things like reasoning, planning
and imagination. Typical system 2 tasks require a sequence
of conscious steps, which also means that they tend
to take more time than system 1 tasks. By this definition,
system 2 abilities are closely related to consciousness.

Cognitive neuroscience has been investigating consciousness
for several decades and a dominant family of theories on which
this paper is anchored are those based on the Global Workspace Theory
~\citep{Baars-1988,Baars-1997,Dehaene+Naccache-2001,Baars-2002,Dehaene-et-al-2017}. This theory posits that we become aware of specific pieces of information which will momentarily form the content of working memory. 
A conscious thought is thus a set of these elements
of which we have become aware, joined together and
made globally available to other computational processes
taking place in the brain at an unconscious level.
Consciousness thus provides a form of bottleneck
for information which has a strong influence on
decision-making (voluntary action), memory (we tend
to very quickly forget what we have not been consciously aware of) and perception (we may be blind to elements
of our sensory input which may distract us from the current
focus of conscious attention).

There are other aspects of consciousness which the global workspace theory does not directly address, such as the notion of self and that of subjective perception, and we do not study them here. Instead, we are interested in
the use of machine learning ideas and experiments as ways
to formalize theories of consciousness (particularly
the global workspace theory), identify advantages which
they can bring to a learning agent (e.g. as a useful
prior for specific aspects of the world), and as
a way to test these theories via machine learning 
experiments measuring for example their effect
on sample efficiency (or the speed of learning) and out-of-distribution
generalization.

\section{Consciousness Prior Theory}

We explain a machine learning framework for
these ideas in more detail below, and place them
in the context of a learning agent with goals (see~\citet{Sutton+Barto-98} 
for basic notions of reinforcement learning).

\subsection{Extracting a Conscious State}
  
Let $x_t$ be the {\bf observation} at time $t$ for a learning agent, and let $h_t$ be the high-level representation
derived from $x_t$ (and from past observed values $\{ x_{t-k}\}$ in the partially observable
case). For example, $h_t$ could be the output of some kind of recurrent neural network (or RNN, with whatever architecture
is appropriate) that reads the sequence
of $x_t$ as input and produces an output $h_t$ at each time step:
\begin{align}
  h_t = F(x_t, h_{t-1})
\end{align}
where we call $F$ the {\bf representation RNN} or encoder and $h_t$ the {\bf unconscious representation state}. We can think of $h_t$ as a very large vector or as a {\bf set} containing all the possible elements which could be brought to consciousness via an attention mechanism.

A core objective for the learner is to learn good representations in $h_t$, which disentangles abstract explanatory
factors, in the sense that there exist a simple transformation of $h_t$ which can select
the information about a single factor (its value or uncertainty about it).
With $h_t$ seen as a set, we can think of each element $e \in h_t$ as one of the
variables over which the learner needs to form a joint distribution in order to make sense of the high-level dependencies. These dependencies do not have to be limited to those between elements in the same $h_t$: they could also relate elements arising at different time steps.

In contrast, we will define the {\bf conscious state} $c_t$ as a very low-dimensional
set which is derived from $h_t$ by a form of attention mechanism applied on $h_t$,
taking into account the previous conscious state and memory as context:
\begin{align}
  c_t = C(h_t, c_{t-1}, m_{t-1}, z_t)
\end{align}
where $z_t$ is a random noise source and $m_t$ is the content of
memory at time $t$. The memory content gets updated by possibly
committing $c_t$ to memory:
\begin{align}
    m_t = M(m_{t-1}, c_t).
\end{align}
We do not explicitly put them in the notation but a realistic agent
would also have goals as part of the context which conditions both
the selection of unconscious items (in $F$) and the update of the
conscious state (in $C$), then seen as a search mechanism.
Also, although we do not explore the architecture of memory mechanisms 
very much here, it is clear
that different kinds of memory mechanisms exist in the brain, starting with short-term
memory from which very recently accessed conscious elements can be retrieved,
as well as longer-term memory, which contains a subset of the elements stored
in short-term memory.
The cognitive interpretation of the above equations
is that the value of $c_t$ is a set of consciously accessed elements and corresponds to the content of a thought one is conscious of at time $t$.
The conscious state $c_t$ is a very small subset of all the information available to us unconsciously, $h_t$, but which has been
brought to our awareness by a particular form of attention which picks several elements or
projections from $h_t$. The function $C$ is the {\bf consciousness process} and because of its
random noise inputs, produces a random choice of the elements on which the attention gets
focused. This is useful if we think of the consciousness process as a tool for exploring interpretations
or plans or to sample predictions about the future or simply
imagined scenarios. We can also think of the consciousness
process as the tool to make a series of associations forming a coherent
argument (for reasoning). It isolates particular high-level abstractions and extracts the information
about each of them (some identifying information and attributes, a value, and uncertainty about it or even the fact that it is observed or not).
This would happen if we think about a single factor, but in general $C$ will aggregate
a few (e.g. a handful) of such factors into a more complex and composed thought.

\subsection{Sparse Factor Graphs}

A factor graph is a way to represent the joint distribution between a set of variables. Let $S=\{V_1, \ldots V_n\}$ be that set and $P(S)$ be their joint distribution. In a factor graph, the joint is represented as a product of potential functions $f_j$, each of which only depends on a subset $S_j \subset S$:
\begin{align}
    P(S) = \frac{\prod_j f_j(S_j)}{Z}
\end{align}
where $Z$ is a normalization constant. We call
each $f_j$ a factor and it creates a direct dependency
between the variables in $S_j$. Indirect
dependencies exist between variables by following
paths in the bipartite graph formed on one hand
with the variables $V_k$ and the factors $f_j$
(each associated with a subset $S_j$ of variables).

Translated in probabilistic terms, the consciousness
prior amounts to the assumption that the factor
graph for the joint distribution between the
elements in the set $h_t$ (or more generally
for the set containing all of the elements in $m_t$
and all those one could think of in the future)
is sparse\footnote{and we probably do not want to represent
that graph explicitly, and instead use conscious attention
to selectively traverse and explore only relevant parts of it,
in the context of given goals}. This is because the cardinality
of all $S_j$'s is small. The motivation for this
assumption comes from observing the structure
of natural language (broken down into phrases,
statements or sentences, each of which involves
very few words) as well as the structure of
formal knowledge representations such as
the sets of facts and rules studied in classical 
symbolic / logic AI or in ontologies and knowledge graphs~\citep{ehrlinger2016towards}.
In addition to being sparse, we believe that
a related assumption can be made: most factors
in the graph describe a strong dependency, i.e.,
one which makes low-entropy predictions (e.g.
about some of the variables in $S_j$ given the
others). Since factor graphs are also generally
understood as energy-based models (the logarithm
of each potential function contributes an additive
term in the energy function corresponding to
the overall joint distribution), we can also
say that each potential function creates a strong
dip in the energy function. Otherwise, they
would not be worth putting in the factor graph.
This is related to the fact that we should think
of this joint distribution as a very rough
approximation of the world built by learning
agents to help them plan, reason, imagine, etc.

An important purpose for the consciousness prior,
from a machine learning point of view, is that
it should help a learner discover an encoder
which captures the kind of high-level variables
which humans talk about when they communicate
with language, since natural language statements
naturally tend to satisfy both the sparsity
requirement (each sentence involves few words)
and the "strong dip" requirement (otherwise
the statement is not worth communicating).
In the quest to discover encoding functions
which disentangle~\citep{Bengio-2009-book,Bengio-Courville-Vincent-TPAMI2013} high-level concepts from each
other, we should see the consciousness prior
as one of many tools to constrain the learner
towards better high-level representations.
Please note in passing that by ``disentangled"
we do not generally mean marginally independent
(that would make all the top-level variables
independent of each other), as in
recent work on variational autoencoders~\citep{higgins2017beta}.
Indeed, notice how natural language concepts
(like say "fork" and "knife") tend to not be
independent of each other, but instead may
be combined to form probable statements
(like "she was eating with her knife and fork").

The analogy with natural language and with
knowledge graphs, ontologies and formal declarative
knowledge also suggests that new potential
functions can be created as needed.
Instead of having a large but fixed set of potential
functions, what we have are mechanisms for
creating new ones which "make sense" according
to observations, reasoning, or imagination.
Instead of enumerating all the possible potential
functions, the brain may have the ability to
instantiate new ones on the fly.
This connects the previous section, which was
about the attention mechanisms for selecting
a small set of variables forming a conscious
thought ($c_t$) with the topic of this section,
which is about the declarative knowledge formed
by the set of potential functions each linking
a few variables together. Whereas the sparse
factor graph constraint is about the underlying
beliefs about the world (when expressed with
the high-level variables), the attention mechanisms
used to build conscious thoughts are part of the
inference mechanisms used to compute efficiently
according to the consciousness prior.

\subsection{Training Objectives}

To capture the assumption that a conscious thought can encapsulate a statement
about the future, we could introduce a {\bf verifier network} which can match
a current representation state $h_t$ with a past conscious state $c_{t-k}$ stored in memory $m_{t-1}$:
\begin{align}
  V(h_t, c_{t-k}) \in \R
\end{align}  
which should be structured so that $V(h_t, c_{t-k})$ indicates the consistency
of $c_{t-k}$ with $h_t$, e.g., estimating the probability of the corresponding
statement being true, given $h_t$.

More generally, we would like to define an objective (or reward) function which embodies the idea
that the attended (conscious) elements are useful, in a way which can
be quantified and optimized, i.e., that the representation RNN
and the attention mechanism which extracts $c_t$ from $h_t$ are trained to optimize this objective
function. This can be in addition to other objectives such as being able to
reconstruct the raw input or any other supervised, RL, or unsupervised
objectives which we probably want to throw in.

There are two distinct mechanisms at play which contribute to map the
high-level state representation to the objective function: (1) the
attention mechanism (e.g. the consciousness RNN) which selects and combines
a few elements from the high-level state representation into a
low-dimensional “conscious sub-state” object (the current content of our
consciousness), and (2) the predictions or actions which are derived from
the sequence of these conscious sub-states. The second mechanism is easy to
grasp and frame in standard ML practice, either in deep learning or RL,
e.g. for supervised or unsupervised or RL tasks. For example, the attention
mechanism could select elements $B$ from the current representation
state and choose to make
a prediction about future elements $A$. Then to improve the quality of the prediction
mechanism we may just want to maximize
$log P(A | B)$ or some proxy for it, e.g., using a
variational auto-encoder~\citep{Kingma+Welling-ICLR2014} objective or a
a conditional GAN~\citep{mirza2014conditional}
if one wants to sample accurately an $A$ from $B$.
Note again that such an objective function is not just used to
learn the mapping from $B$ to $A$ (or to probabilities over the space of $A$
values), but also drives the learning of the representation function
itself, i.e., is back-propagated into the representation RNN). However,
this part of the objective function (e.g. predictive value, computed
by $V$ above) is not
sufficient and in fact is not appropriate to train the attention mechanism
itself (which variables $A$ and $B$ should be selected?). Indeed, if that was
the driving objective for attention, the learner would always pick a pair $(A,B)$ such that $A$ is trivially predictable from $B$ (and there are such aspects of reality which
are trivially predictable yet do not help us to further understand the
world and make sense of it or achieve our goals). It remains an open
question what other objectives would be appropriate for learning how to
attend to the most useful elements, but ultimately we should be able to use
the actual RL reward of the learning agent using $c_t$ for taking decisions. Some form of mutual information, entropy or diversity may
be needed so that the attention mechanism is stochastic and can choose a
very diverse set of possible attended elements, so as to cover widely the possible variables
$A$ on which a prediction is made, i.e., the entropy of $(A,B)$ pairs.

\subsection{Naming Variables and Indirection}

Content-based soft-attention or hard-attention mechanisms~\citep{Bahdanau-et-al-ICLR2015-small,Xu-et-al-ICML2015} extract a value from a set of element by taking a convex weighted sum of values from an input set of values. Those weights are the attention weights and they are computed by an attention mechanism which gives a larger weight on the element with the  most appropriate "key", according to some context.

In standard neural networks without attention, a neuron $i$ is identified by its position in its layer and the signal it sends to some other neuron $j$ downstream does not need to be identified as coming from $i$. However, when attention mechanisms such as described above are used to provide an input value to $j$, the input could come from any of the elements over which attention is making a selection. Depending on the computation performed, it could thus be useful for downstream layers with attention mechanisms selecting their input to receive not just the weighted (selected) value but also information about the source of the information. We can think of that information as a variable name (and possibly other attributes which we can interpret as variable type), which complement the variable value. The idea of (key,value) pairs was used in memory augmented neural networks~\citep{Weston2014,Graves2014}, although it is not clear if a distinction between keys and values exists in the brain, or if a general auto-associative mechanism is used instead.

When elements from the unconscious state $h_t$ are selected to enter the conscious state $c_t$ using content-based soft-attention~\citep{Bahdanau-et-al-ICLR2015-small}, it is not just a value which should be copied but also some "key" which identifies the origin of that value. Modern attention-based deep learning architectures such as Transformers~\citep{vaswani2017attention} bind (key,value) pairs together precisely for that purpose. For example, the kind of verifier network discussed above needs to associate a (key,prediction) pair made in the past with a (key,realization) pair observed later. The key thus acts like a name and provides a form of indirection or reference.  If the key and value were mixed up and
the predicted value differs substantially from the observed value, a simple
associative process might miss the opportunity to match these and thus provide
a strong training signal (to correct the predictor).
Another reason to represent keys separately from values is that the keys can be used to represent a form of type information, to help match the expected argument type of a downstream computation with an appropriate element selected by an attention mechanism. This is important in order to obtain systematic generalization~\citep{lake2017generalization} and combinatorial properties omnipresent in natural language, making it easier to combine different pieces of neural hardware together dynamically, with keys being used to decide which information should be routed where.
We could thus see the conscious state as a bottleneck to route such information
across many different modules.

\subsection{Connection to Language and Symbolic Knowledge Representation}

We hypothesize that conscious
processing of the kind described
above could thus help the brain
(and future machine learning systems) achieve better
systematic generalization
and combine concepts in
fluent and combinatorial ways.
The fact that we define
consciousness in terms of
verbal reporting may be important
to note here.
All this indeed suggests
that there is a fairly simple transformation of conscious
states into
natural language sentences.
Conversely, an externally provided sentence (heard or read) could also elicit an associated conscious state, although
we postulate that the conscious state is generally a richer object than the uttered
sentence, i.e., mapping from conscious states to sentences loses information
(think about visual imagery, or artistic expression, which are difficult to put in words),
and the same sentence could thus be interpreted differently depending on context and
the particulars of the agent who reads that sentence.
Formally, we could use another RNN to map a conscious state to an utterance $u_t$:
\begin{align}
  u_t = U(c_t, u_{t-1}).
\end{align}  
A learning agent which uses language could thus benefit from an additional
regularization effect putting
pressure on the encoder: the set of currently consciously attended elements should have a direct two-way mapping
with natural
language utterances
which may be uttered by other agents, such as a human teacher.
This would act as a weak form
of supervision for the concepts
produced by the encoder.
A sentence focuses on just a handful of elements and
concepts, unlike our full internal state. This imposes soft constraints
on the representation function in that its individual elements or
dimensions are more likely to correspond to concepts which can typically be
expressed by a single word or phrase.
Based on these arguments, it is reasonable to hypothesize that language may
actually help humans build sharper internal representations (which are
better disentangled) as well as facilitate learning -- see the
arguments around curriculum learning~\citep{Bengio+al-2009-small}
and cultural learning~\citep{bengio-2014-cultural} --
and enable collaborative task-solving.

Along the same line, this research opens the door to the possibility of
better connecting deep learning with classical symbolic AI and cognitive
science, and move deep learning from perception (where it currently shines) to
higher-level cognition and knowledge representation (where many questions
remain open). For example, declarative knowledge is classically represented by facts
and rules: each of them is a very sharp statement (true with high probability) about reality involving
just a few concepts. Such a nugget of information or knowledge seems to fit
well as a conscious state. Combining such conscious states
sequentially in order to make more complex predictions and inferences or
actions is basically what reasoning is about. However, pasting symbolic logic
computations on top of a deep learning encoder might not succeed for several
reasons. This would lose the ability manipulate uncertainty as well as
represent the context-dependent effect of goals and background knowledge
which deep learning with content-based attention can provide, in addition
to the ability to improve generalization through distributed representations.
Instead, we envision extensions of deep learning based on attention
that implement conscious processing functionalities associated with 
system 2 tasks in humans.
Progress in this direction would also address the often expressed
concern about obtaining explanations from deep nets, since the approach
proposed here would make it easier for a trained agent to communicate
verbally its high-level state.

\section{Considerations for Experimenting with the Consciousness Prior}

Because this is a novel theory which may be developped in many different ways,
it is important to start with simple toy experiments allowing one to test
and evaluate qualitatively different approaches, such that the turnaround
time for each experiment is very short and the analysis of the representations
learned very easy (because we already have a preconceived idea of what
concepts would be the most appropriate to disentangle).

Although working with natural language input would be likely to help the agent
learn better and more abstract representations, it might be better to start
with experiments with no linguistic input, to make sure that it is the training
objective and the training framework alone which are leading to the discovery
of the appropriate high-level concepts. For example, learning some form
of intuitive physics is done by babies without the need for linguistic
guidance. Similarly, although the consciousness prior could be used in
supervised learning or task-oriented RL, testing its ability alone to discover
high-level abstractions would be best done in the context of unsupervised
RL, e.g., using an intrinsic reward which favours the discovery of how
the environment works.

It would be more interesting for the learning task to involve meaningful abstractions
which have a high predictive power. For example, consider predicting whether a pile of blocks
will fall on or off a table. It involves a high-level discrete outcome which can be
predicted easily, even if the details of where the blocks will fall is very difficult
even for humans to predict. In that case, predicting the future at the pixel level
would be extremely difficult because future states have high entropy, with a highly
multi-modal distribution. However, some aspects of the future may have low entropy.
If in addition, these aspects have a big impact on predicting what will come next
(or on taking the right decisions now), then the consciousness prior should be
very useful.

\section*{Acknowledgements}

The author wants to thank Philippe Beaudoin, Gerry (Tong) Che, William Fedus, Devon
Hjelm and Anirudh Goyal for preliminary discussions about the consciousness prior,
as well as funding from NSERC, CIFAR, the Canada Research Chairs, and the Open Philanthropy
Project.

\bibliographystyle{plainnat}
\bibliography{strings,aigaion,ML,ML+,conscious}

\end{document}